\documentclass[12pt,twoside]{article}

\usepackage{
latexsym,
graphicx,
times,
fancyhdr,
amssymb,
amsmath,
stmaryrd,
fleqn,
subfigure,
psfrag
}
\newtheorem{definition}{Definition}

\begin{document}
\thispagestyle{empty}

\title{\vspace{-6ex}
{\bf Belief Calculus}
}
\author{
{\normalsize \bf Audun J{\o}sang}
\vspace{2ex}\\
{\normalsize Queensland University of Technology}
{GPO Box 2434, Brisbane Qld 4001, Australia}\\
{\normalsize tel:+61-7-3864 2960, fax:+61-7-3864 1214}\\
{\normalsize email: a.josang@qut.edu.au}}
\date{}
\maketitle
\section*{\large Abstract}
{\em In Dempster-Shafer belief theory, general beliefs are expressed
as belief mass distribution functions over frames of discernment. In
Subjective Logic beliefs are expressed as belief mass distribution
functions over binary frames of discernment. Belief representations in
Subjective Logic, which are called opinions, also contain a base rate
parameter which express the {\em a priori} belief in the absence of
evidence. Philosophically, beliefs are quantitative representations
of evidence as perceived by humans or by other intelligent agents. The
basic operators of classical probability calculus, such as addition
and multiplication, can be applied to opinions, thereby making belief
calculus practical. Through the equivalence between opinions and
Beta probability density functions, this also provides a calculus
for Beta probability density functions. This article explains the
basic elements of belief calculus. }

\section{Introduction}

Belief theory has its origin in a model for upper and lower
probabilities proposed by Dempster in 1960. Shafer later proposed a
model for expressing beliefs \cite{Sha1976}. The main idea behind
belief theory is to abandon the additivity principle of probability
theory, i.e. that the sum of probabilities on all pairwise exclusive
possibilities must add up to one. Instead belief theory gives
observers the ability to assign so-called belief mass to any subset of
the frame of discernment, i.e. non-exclusive possibilities including
the whole frame itself. The main advantage of this approach is that
ignorance, i.e. the lack of information, can be explicitly expressed
e.g. by assigning belief mass to the whole frame. Shafer's book
\cite{Sha1976} describes many aspects of belief theory, but the two
main elements are 1) a flexible way of expressing beliefs, and 2) a
method for combining beliefs, commonly known as Dempster's Rule.

\section{Representing Opinions}
\label{sec:uncertain-beliefs}

Belief representation in classic belief theory\cite{Sha1976} is based
on an exhaustive set of mutually exclusive atomic states which is
called the {\em frame of discernment} $\Theta$.  The power set
$2^{\Theta}$ is the set of all sub-states of $\Theta$.  A
bba (basic belief assignment\footnote{Called {\em basic probability assignment} in
\cite{Sha1976}.}) is a belief distribution
function $m_{\Theta}$ mapping $2^{\Theta}$ to $[0,1]$ such that
\begin{equation}
\sum_{x \subseteq {\Theta}} m_{\Theta}(x) = 1\;, \;\mbox{ where }\; m_{\Theta}(\emptyset) = 0\;.
\end{equation}
The bba distributes a total belief mass of 1 amongst the subsets of
$\Theta$ such that the belief mass for each subset is positive or
zero. Each subset $x \subseteq \Theta$ such that $m_{\Theta}(x)>0$ is
called a focal element of $m_{\Theta}$. In the case of total
ignorance, $m_{\Theta}(\Theta) = 1$, and $m_{\Theta}$ is called a {\em
vacuous} bba. In case all focal elements are atoms (i.e. one-element
subsets of $\Theta$) then we speak about {\em Bayesian} bba. A {\em
dogmatic} bba is when $m_{\Theta}(\Theta) = 0$ \cite{Sme1988}. Let us
note that, trivially, every Bayesian bba is dogmatic.

In addition, we also define the Dirichlet bba, and its cluster variant,
as follows.

\begin{definition}[Dirichlet bba]\hspace*{1.5mm}
\label{def:Dirichlet-belief-function}
A bba where the possible focal elements are $\Theta$ and/or singletons of
$\Theta$, is called a Dirichlet belief
distribution function.
\end{definition}

\begin{definition}[Cluster Dirichlet bba]\hspace*{1.5mm}
\label{def:coarse-Dirichlet-belief-function}
A bba where the only focal elements are $\Theta$ and/or mutually
exclusive subsets of $\Theta$ (singletons or clusters of singletons),
is called a cluster Dirichlet belief distribution function.
\end{definition}

It can be noted that Bayesian bbas are a special case of Dirichlet
bbas.  

The Dempster-Shafer theory \cite{Sha1976} defines a belief function
$b(x)$. The probability transformation \cite{DP1982}\footnote{Also
known as the pignistic transformation \cite{Sme2005-IJAR,SK1994}}
projects a bba onto a probability expectation value denoted by
$\mbox{E}(x)$. In addition, subjective logic \cite{Jos2001-IJUFKS}
defines a disbelief function $d(x)$, an uncertainty function $u(x)$,
and a base rate function\footnote{Called {\em relative atomicity} in
\cite{Jos2001-IJUFKS}.}  $a(x)$, defined as follows:

\begin{alignat}{2}
b(x) &= \sum_{\emptyset \neq y \subseteq x} m(y) \qquad
&\forall \; x \subseteq {\Theta}\;,
\label{equa:bel}\\
d(x) &= \sum_{y \cap x = \emptyset} m(y) \qquad
&\forall \; x \subseteq {\Theta}\;,
\label{equa:dis} \\
u(x) &= \sum_{\scriptsize \scriptsize \begin{array}{cc}{y} \cap {x} \neq \emptyset\\
{y} \not\subseteq {x}\end{array}} m(y) \qquad
&\forall \; x \subseteq {\Theta}\;,
\label{equa:unc}\\
a(x) &= {{|x|} / {|\Theta}|} \qquad
&\forall \; x \subseteq {\Theta}\;,
\label{equa:ato}\\
\mathrm{E}(x) &= \sum_{y \subseteq \Theta} m_{\Theta}(y)\frac{|x\cap y|}{|y|} \qquad
&\forall \; x \subseteq {\Theta}\;.
\label{equa:exp}
\end{alignat}

In case $|\Theta| > 2$, coarsening is necessary in order to apply
subjective logic operators. Choose $x \subseteq {\Theta}$, and let
$\overline{x}$ be the complement of $x$ in $\Theta$, then $X =
\{x,\overline{x}\}$ is a binary frame, and $x$ is called that {\em
target} of the coarsening. The coarsened bba on $X$ can consist of the
three belief masses $b_{x} = m_{X}(x)$, $d_{x} = m_{X}(\overline{x})$
and $u_{x} = m_{X}(X)$, called {\em belief}, {\em disbelief} and {\em
uncertainty} respectively. Coarsened belief masses can be computed
e.g. with simple or normal coarsening as defined in
\cite{Jos2001-IJUFKS,JM2004-IJAR}, or with {\em smooth
coarsening} defined next.

\begin{definition}[Smooth Coarsening]
\label{def:smooth-coarsening}
Let $\Theta$ be a frame of discernment and let $b(x)$, $d(x)$, $u(x)$
and $a(x)$ be the belief, disbelief, uncertainty and base rate
functions of the target state ${x} \subseteq \Theta$, with probability
expectation value $\mathrm{E}(x)$. Let $X = \{x,\overline{x}\}$ be the
binary frame of discernment, where $\overline{x}$ is the complement of
$x$ in $\Theta$. Smooth coarsening of $\Theta$ into $X$ produces the
corresponding belief, disbelief, uncertainty and base rate functions
$b_{x}$, $d_{x}$, $u_{x}$ and $a_{x}$ defined by:\\
\[
\begin{array}{l}
\mbox{For }\;\mathrm{E}(x) \leq b(x) + a(x)u(x)\;:\\
\hline\\
\left\{
\begin{array}{ll}
b_{x} &= \frac{\mathrm{E}(x)b(x)}{b(x) + a(x)u(x)}\\\\

d_{x} &= 1 -\frac{\mathrm{E}(x)(b(x) + u(x))}{b(x) + a(x)u(x)}\\\\

u_{x} &= \frac{\mathrm{E}(x)u(x)}{b(x) + a(x)u(x)}\\\\

a_{x} &= a(x)\;,

\end{array}
\right.
\\\\
\mbox{for }\;\mathrm{E}(x) > b(x) + a(x)u(x)\;:\\
\hline\\
\left\{
\begin{array}{ll}
b_{x} &= 1 - \frac{(1 - \mathrm{E}(x))(d(x) + u(x))}{1 - b(x) - a(x)u(x)}\\\\

d_{x} &= \frac{(1-\mathrm{E}(x))d(x)}{1 - b(x) - a(x)u(x)}\\\\

u_{x} &= \frac{(1-\mathrm{E}(x))u(x)}{1 - b(x) - a(x)u(x)}\\\\

a_{x} &= a(x)\;.

\end{array}
\right.
\end{array}
\]
\end{definition}

In case the target element for the coarsening is a focal element of a
(cluster) Dirichlet belief distribution function, $\mathrm{E}(x) =
b(x) + a(x)u(x)$ holds, and the coarsening can be described as stable.

\begin{definition}[Stable Coarsening]
\label{def:stable-coarsening}
Let $\Theta$ be a frame of discernment, and let $x \in \Theta$ be the
target element for the coarsening. Let the belief distribution function be a
Dirichlet (cluster or not) bba such that the target element
is also a focal element. Then the stable coarsening of $\Theta$
into $X$ produces the belief, disbelief and uncertainty functions
$b_{x}$, $d_{x}$ and $u_{x}$ defined by:
\[
\begin{array}{lll}
b_{x} = b(x)\;,\;\;&d_{x} = d(x)\;,\;\;&u_{x} = u(x)\;.
\end{array}
\]
\end{definition}

It can be shown that $b_x, d_x, u_x, a_x \in [0,1]$, and that Eq.(\ref{eq:E01}) and  Eq.(\ref{eq:bipolar-expectation}) hold.
\begin{equation}
\label{eq:E01}
b_x+d_x+u_x = 1\;.\;\;\;\;\;\;\mbox{(Additivity)}
\end{equation}

\begin{equation}
\label{eq:bipolar-expectation}
\mathrm{E}(x) = b_x + a_x u_x\;.\;\;\;\;\;\mbox{(Expectation)} 
\end{equation}

It can be noticed that non-stable coarsening requires an
adjustment of the belief, disbelief and uncertainty functions in
general, and this can result in inconsistency when applying the
consensus operator. The purpose of this paper is to present a method
to rectify this problem. This will be explained in further detail in
the sections below.

The ordered quadruple $\omega_x = (b_x,\; d_x,\; u_x,\; a_x)$, called
the opinion about $x$, is equivalent to a bba on a binary frame of
discernment $X$, with an additional base rate parameter $a_x$ which
can carry information about the relative size of $x$ in $\Theta$.

The opinion space can be mapped into the interior of an equal-sided
triangle, where the the relative distance towards the bottom right,
bottom left and the top corners represent belief, disbelief and
uncertainty functions respectively. For an arbitrary opinion $\omega_x
= (b_x,d_x,u_x,a_x)$, the three parameters $b_x$, $d_x$ and $u_x$
determine the position of the opinion point in the triangle. The base
line is the {\em probability axis}. The base rate value can be
indicated as a point on the probability axis.

\begin{figure}[h]
\begin{center}
\includegraphics[scale=0.45]{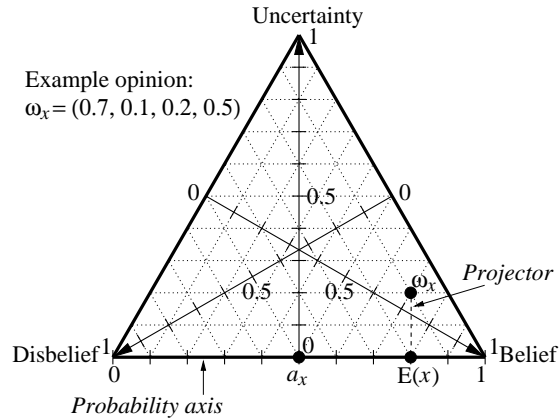}
\caption{Opinion triangle with example opinion}
\label{fig:tri-7-1}
\end{center}
\end{figure}

Fig.\ref{fig:tri-7-1} illustrates an example
opinion about $x$ with the value
$\omega_{x}=(0.7,\;0.1,\;0.2,\;0.5)$.

The {\em projector} going through the opinion point, parallel to the
line that joins the uncertainty corner and the base rate point,
determines the probability expectation value $\mathrm{E}(x) = b_x + a_x u_x$.

Although an opinion has 4 parameters, it only has 3 degrees of freedom
because the three components $b_{x}$, $d_{x}$ and $u_{x}$ are
dependent through Eq.(\ref{eq:E01}). As such they represent the
traditional $\mbox{\it Bel}(x)$ (Belief) and $\mbox{\it Pl}(x)$
(Plausibility) pair of Shaferian belief theory through the
correspondence $\mbox{\it Bel}(x) = b(x)$ and $\mbox{\it Pl}(x) = b(x)
+ u(x)$. The disbelief function $d(x)$ is the same as {\em doubt} of
${x}$ in Shafer's book. However, `disbelief' seems to be a better term
because the case when it is certain that $x$ is false, is better
described by `total disbelief' than by `total doubt'.

The reason why a redundant parameter is kept in the opinion
representation is that it allows for more compact expressions of
opinion operators, see
e.g. \cite{Jos2001-IJUFKS,Jos2002-IPMU,JM2004-IJAR,JPD2005-ECSQARU,JPM2006-iTrust}.

Various visualisations of opinions are possible to facilitate human
interpretation. For this, see
http://sky.fit.qut.edu.au/{$\sim$}josang/sl/demo/BV.html

\section{Mapping Opinions to Beta PDFs}

The beta-family of distributions is a continuous
family of distribution functions indexed by the two parameters
$\alpha$ and $\beta$. The beta distribution $\mbox{beta}(\alpha,\beta)$
can be expressed using the gamma function $\Gamma$ as:
\begin{equation}
\label{eq:F01}
\begin{array}{l}
\mbox{beta}(\alpha,\beta) = {{\Gamma(\alpha  + \beta)} \over
{\Gamma(\alpha)\Gamma(\beta)}} {P}^{\alpha-1}(1\!-\!{P})^{\beta-1}\;,\\
\mbox{ where } 0\leq{P}\leq1,\;\alpha>0,\;\beta>0\;
\end{array}
\end{equation}
with the restriction that the probability ${P} \neq 0$ if $\alpha <
1$, and ${P} \neq 1$ if $\beta < 1$. The probability expectation value
of the beta distribution is given by: 
\begin{equation}
\mathrm{E}(P) = \frac{\alpha}{\alpha + \beta}
\end{equation}\;,
where $P$ is the random variable corresponding to
the probability.

It can be observed that the beta PDF has two degrees of freedom
whereas opinions have three degrees of freedom as explained in
Sec.\ref{sec:uncertain-beliefs}. In order to define a bijective
mapping between opinions and beta PDFs, we will augment the beta PDF
expression with 1 additional parameter representing the prior, so
that it also gets 3 degrees of freedom.

The $\alpha$ parameter represents the amount of evidence in favour a
given outcome or statement, and the $\beta$ parameter represents the
amount of evidence against the same outcome or statement.  With a
given state space, it is possible to express the {\em a priori} PDF using a
base rate parameter in addition to the evidence parameters.

The beta PDF parameters with the prior base rate $a$ included can be
defined as \cite{Jos2001-IJUFKS}:
\begin{equation}
\label{eq:beta-correspondence}
\left\{
\begin{array}{l}
\alpha = r + 2a,\\
\beta  = s + 2(1-a)\;,
\end{array}
\right.
\end{equation}
where $a$ represents the {\em a priori} base rate, $r$ represents the
amount of positive evidence, and $s$ represents the amount of negative
evidence.

We define the {\em augmented beta PDF}, denoted $\varphi(r,s,a)$, with 3 parameters
as:

\begin{equation}
\label{eq:augmented-beta-PDF}
\varphi(r,s,a) = \mbox{beta}(\alpha,\beta)\;, \mbox{given Eq.(\ref{eq:beta-correspondence})}.
\end{equation}

This augmented beta distribution function distinguishes between the
{\em a priori} base rate $a$, and the {\em a posteriori} observed
evidence $(r,s)$.

The probability expectation value
of the augmented Beta distribution is given by: 
\begin{equation}
\label{eq:augmented-beta-exp}
\mathrm{E}(P) = \frac{r + 2a}{r + s + 2}
\end{equation}\;,
where $P$ is the random variable corresponding to
the probability.

\begin{figure}[h]
{\centering
\includegraphics[scale=0.5]{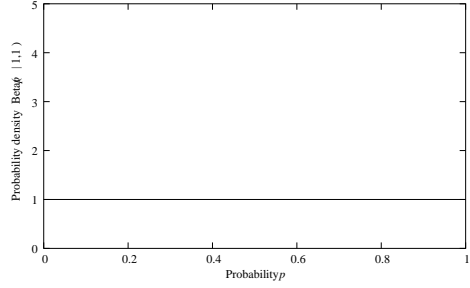}
\caption{{\em A priori} $\varphi(0,0,\frac{1}{2})$}
\label{fig:beta-1-1}
}
\end{figure}

For example, when an urn contains unknown proportions of red and black
balls, the likelihood of picking a red ball is not expected to be
greater or less than that of picking a black ball, so the {\it a
priori} probability of picking a red ball is $a=0.5$, and the {\it a
priori} augmented beta distribution is $\varphi(0,0,\frac{1}{2}) =
\mbox{beta}(1,1)$ as illustrated in Fig.\ref{fig:beta-1-1}.

\begin{figure}[h]
{\centering 
\includegraphics[scale=0.5]{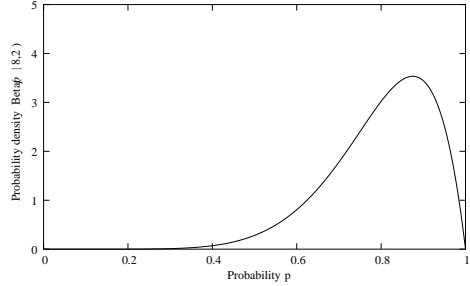}
\caption{Updated $\varphi(7,1,\frac{1}{2})$}
\label{fig:beta-8-2}
}
\end{figure}

Assume that an observer picks 8 balls, of which 7 turn out to be red
and only one turns out to be black. The updated augmented beta distribution of
the outcome of picking red balls is $
\varphi(7,1,\frac{1}{2}) = \mbox{beta}(8,2)$ which is illustrated in
Fig.\ref{fig:beta-8-2}.

The expression for augmented beta PDFs has 3 degrees of freedom and
allows a bijective mapping to opinions \cite{Jos2001-IJUFKS}, defined
as:\\

\begin{equation}
\begin{array}{lll}
\!\!\!\!\!\!\!\!\!\!\!\!\!\!\!\!\left\{
\begin{array}{l}
r = 2b_x/u_x\\
s = 2d_x/u_x\\
1 = b_{x}\! +\! d_{x}\! +\! u_{x}\\
a = a_x\;,
\end{array}
\right.
\!\!\!\!\!\!\!&\Leftrightarrow
\!\!\!\!\!&\left\{
\begin{array}{l}
b_{x} = r/(r\! +\! s\! +\! 2)\\
d_{x} = s/(r\! +\! s\! +\! 2)\\
u_{x} = 2/(r\! +\! s\! +\! 2)\\
a_{x} = a\;.
\end{array}
\right.
\end{array}
\label{eq:bijective-mapping}
\end{equation}

It can be noted that under this correspondence, the example opinion of
Fig.\ref{fig:tri-7-1} and the beta distribution of
Fig.\ref{fig:beta-8-2} are equivalent.

\section{Mapping Opinions to Basic Probability Vectors}

Both the opinion and the augmented Beta representation have the
inconvenience that they do not explicitly express the probability
expectation value. Although simple to compute with either
Eq.\ref{eq:bipolar-expectation} or Eq.\ref{eq:augmented-beta-exp}, it
represent a barrier for quick and intuitive interpretation by
humans. We therefore propose a representation in the form of a Basic
Probability Vector with three degrees of freedom that explicitly
expresses the probability expectation value.

\begin{definition}{Basic Probability Vector}
Let $\Theta$ be a frame of discernment where $x \subset \Theta$ , and
let $e$ be the subjective probability expectation value of $x$ as seen
by an observer $A$. Let $u$ be the uncertainty of $e$, equivalent to
the uncertainty defined for opinions. Let $a$ be the base rate of $x$
in $\Theta$, equivalent of the base rate defined for opinions. The
Basic Probability Vector, denoted by $\pi^{A}_{x}$, is defined as
$\pi^{A}_{x} = (e,u,a)$
\end{definition}

In this context, the probability $p$ is simple to interpret. In case
$u=0$, then $e$ is a frequentist probability. In case $u=1$, then $e =
a$, i.e. $a$ is the probability expectation value of an augmented Beta
probability density function where $r=s=0$.

The equivalence between opinions and basic probability vectors is
defined below. The uncertainty and base rate parameters are identical
in both representations, and their mapping is therefore not needed.

\begin{equation}
\begin{array}{lll}
\left\{\begin{array}{l}
e=b+au\\
b+d+u=1
\end{array}
\right.
&\Leftrightarrow
&\left\{
\begin{array}{ll}
b=e-au\\
d=1-e - u(1-a)
\end{array}
\right.
\end{array}
\end{equation}

\section{Addition and Subtraction of Opinions}
Addition of beliefs corresponds to determining the probability of the
disjunction of two mutually exclusive subsets in a frame of
discernment, given the probabilities of the original addend subsets. This is illustrated in Fig.\ref{fig:addition} below.

\begin{figure}[h]
\begin{center}
\includegraphics[scale=0.45, angle=-90]{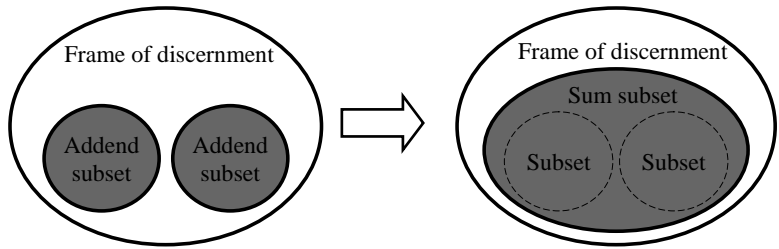}
\caption{Addition principle}
\label{fig:addition}
\end{center}
\end{figure}

Subtraction of beliefs correspond to determining the probability of
the difference of two subsets, given the probabilities of the original
minuend and subtrahend subsets, where the subset of the minuend
contains the subset of the subtrahend.

\begin{figure}[h]
\begin{center}
\includegraphics[scale=0.45, angle=-90]{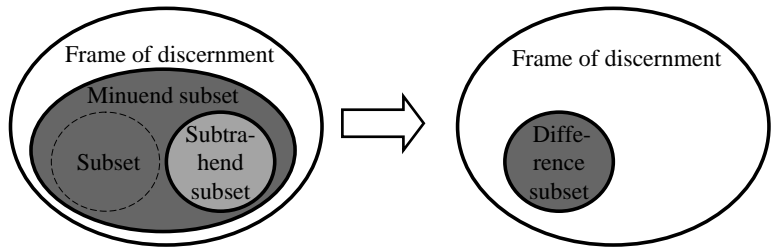}
\caption{Subtraction principle}
\label{fig:subtraction}
\end{center}
\end{figure}

\subsection{Addition of Opinions}
\label{addition}

The addition of opinions in subjective logic is a binary operator that 
takes opinions in a single frame of discernment about two mutually
exclusive alternatives ({\em i.e.} two disjoint subsets of the frame
of discernment), and outputs an opinion about the union of the subsets.
Let the two sets be denoted by $x$ and $y$, so $x$ and $y$ are subsets 
of $\Theta$ such that $x \cap y = \emptyset$, then we are interested in
the opinion about $x \cup y$, given
opinions about $x$ and $y$.  Since $x$ and $y$ are mutually exclusive, then
it is to be expected that $b_x \leq d_y$ and $b_y \leq d_x$ as belief in
either necessarily requires disbelief in the other.  Since $x$ and $y$ are
mutually exclusive, the belief in the union must be able to account for
both the belief in $x$ and the belief in $y$, so that the first option that
can be considered is $b_{x \cup y} = b_x + b_y$, so that $b_{x \cup y}$ has
been partitioned exactly into the two possibilities.  In order to calculate
the atomicity of $x \cup y$, it is merely necessary to point out that
\begin{equation*}
\begin{array}{ll}
a(x \cup y) &= \frac{|x \cup y|}{|\Theta|} = \frac{|x| + |y|}{|\Theta|}
= \frac{|x|}{|\Theta|} + \frac{|y|}{|\Theta|}\\\\
&= a(x) + a(y),
\end{array}
\end{equation*}
as $|x \cup y| = |x| + |y|$ since $x$ and $y$ are disjoint.  Therefore 
the atomicity of $x \cup y$ is given by $a_{x \cup y} = a_x + a_y$.  
Similarly, the expectation value of $x \cup y$ must be given by
\begin{equation*}
E(x \cup y) = E(x) + E(y),
\end{equation*}
since the probabilities of $x$ and $y$ are summed to calculate the
probability of $x \cup y$ in probability calculus, and expectation value
is linear on random variables ({\em i.e.} if $U$ and $V$ are random variables
and $\mu$ and $\nu$ are real numbers, then $E[\mu U + \nu V] = \mu E[U] + \nu
E[V]$).  This gives us sufficient information to be able to calculate the
opinion about $x \cup y$, with the result that
\begin{align}
\label{addition-formula}
b_{x \cup y} &= b_x + b_y,\\
d_{x \cup y} &= \frac{a_x (d_x - b_y) + a_y (d_y - b_x)}{a_x + a_y},\\
u_{x \cup y} &= \frac{a_x u_x + a_y u_y}{a_x + a_y},\\
\label{addition-formula-end}
a_{x \cup y} &= a_x + a_y.
\end{align}

By using the symbol ``+'' to denote the addition operator for opinions, we can write:
\begin{equation}
\omega_x + \omega_y = \omega_{x \cup y}\;.
\end{equation}

Note that the uncertainty of $x \cup y$ is the weighted average of the
uncertainties of $x$ and $y$, and the disbelief in $x \cup y$ is the weighted 
average of what is left in each case when the belief in one is subtracted from
the disbelief in the other (when the amount of probability that neither is true
is estimated), {\em i.e.} the disbelief in $x \cup y$ is the weighted average
of the two estimates of the probability that the system is in neither the 
state $x$ nor the state $y$.

In the case where $u_x = u_y$ ({\em i.e.} $x$ and $y$ have the same 
uncertainty), then
\begin{align}
\label{addition-unc}
b_{x \cup y} &= b_x + b_y,\\
d_{x \cup y} &= d_x - b_y = d_y - b_x,\\
u_{x \cup y} &= u_x = u_y,\\
\label{addition-unc-end}
a_{x \cup y} &= a_x + a_y,
\end{align}
so that the uncertainty in $x \cup y$ is equal to the common value of the 
uncertainty of $x$ and the uncertainty of $y$.  The disbelief $d_{x \cup y}$
in $x \cup y$ is equal to the result when the belief in either is subtracted
from the disbelief in the other.  The fact that $d_{x \cup y} = d_x - b_y =
d_y - b_x$ can be justified by the same sort of arguments that were used
to justify $b_{x \cup y} = b_x + b_y$.  Since the states $x$ and $y$ are 
mutually exclusive, then belief in $x$ necessitates disbelief in $y$, so
that disbelief in $y$ can be partitioned into that part which corresponds
to belief in $x$ (of magnitude $b_x$) and that part which corresponds to
belief in neither $x$ nor $y$ (of magnitude $d_y - b_x$).  Since the
latter part of the disbelief in $y$ (of magnitude $d_y - b_x$) corresponds
to belief in neither $x$ nor $y$, then it corresponds to disbelief in
$x \cup y$, and so it is reasonable that $d_{x \cup y} = d_y - b_x$.

Various considerations make it natural that disjoint states in the
frame of discernment should have the same uncertainty (and in fact, by
the same considerations, all states in the frame of discernment should
have the same uncertainty), especially if the {\em a priori} opinions
are being formed, so that the uncertainty is equal to 1 to reflect the
complete ignorance about the probabilities, and through updating due
to evidence, where the uncertainty is dependent only on the amount of
evidence that has been gathered.  But this does not mean that the
formulae given in Equations
\ref{addition-formula}-\ref{addition-formula-end} should be
disregarded.  As noted in the Introduction, there are various
operators which can be applied to opinions, and if the results of two
such calculations result in opinions about two disjoint states in the
same frame of discernment, the likelihood is that the opinions about
the two states have unequal uncertainties, and the formulae that have
to be used to calculate the sum of the opinions are Equations
\ref{addition-formula}-\ref{addition-formula-end}, rather than the
simpler formulae in Equations
\ref{addition-unc}-\ref{addition-unc-end}.  This is certainly the
case, for example, when $x_1$ and $x_2$ are states in one frame of
discernment, and $y_1$ and $y_2$ are states in another frame of
discernment, and either $x_1$ and $x_2$ are disjoint states, or $y_1$
and $y_2$ are disjoint states.  If the simple products
\cite{JM2004-IJAR} of opinions ($x_1 \land y_1$ and $x_2 \land y_2$)
are determined, then they will almost certainly not have the same
uncertainty.  Similarly, if the normal products \cite{JM2004-IJAR} are
determined, then they will almost certainly not have the same
uncertainty.

\subsection{Subtraction of Opinions}

The inverse operation to addition is subtraction.  Since addition of
opinions yields the opinion about $x \cup y$ from the opinions about
disjoint subsets of the frame of discernment, then the difference between the
opinions about $x$ and $y$ ({\em i.e.} the opinion about $x \backslash y$) can
only be
defined if $y \subseteq x$ where $x$ and $y$ are being treated as subsets of
$\Theta$, the frame of discernment, {\em i.e.} the system must be in the state
$x$ whenever it is in the state $y$.  The opinion about $x \backslash y$ is
the opinion
about that state which consists exactly of the atomic states of $x$ which are
not also atomic states of $y$, {\em i.e.} the opinion about the state in which
the system is in the state $x$ but not in the state $y$.  Since $y \subseteq
x$, then belief in $y$ requires belief in $x$, and disbelief in $x$ requires
disbelief in $y$, so that it is reasonable to require that $b_x \geq b_y$ and
$d_x \leq d_y$.  Since $y \subseteq x$, then $|y| \leq |x|$, so that $a(y)
\leq a(x)$, and so $a_y \leq a_x$.  The opinion about $x \backslash y$ is
given by
\begin{align}
b_{x \backslash y} &= b_x - b_y,\\
d_{x \backslash y} &= \frac{a_x (d_x + b_y) - a_y (1 + b_y - b_x - u_y)}
{a_x - a_y},\\
u_{x \backslash y} &= \frac{a_x u_x - a_y u_y}{a_x - a_y},\\
a_{x \backslash y} &= a_x - a_y.
\end{align}
Since $u_{x \backslash y}$ should be nonnegative, then this requires that
$a_y u_y \le a_x u_x$, and since $d_{x \backslash y}$ should be nonnegative,
then this requires that $a_x (d_x + b_y) \geq a_y (1 + b_y - b_x - u_y)$.

By using the symbol ``-'' to denote the subtraction operator for opinions, we can write:
\begin{equation}
\omega_x - \omega_y = \omega_{x \backslash y}\;.
\end{equation}

In the case where $u_x = u_y$ ({\em i.e.} $x$ and $y$ have the same 
uncertainty), then
\begin{align}
b_{x \backslash y} &= b_x - b_y,\\
d_{x \backslash y} &= d_x + b_y,\\
u_{x \backslash y} &= u_x = u_y,\\
a_{x \backslash y} &= a_x - a_y,
\end{align}
so that the uncertainty in $x \backslash y$ is equal to the common value of the 
uncertainty of $x$ and the uncertainty of $y$.  The belief $b_{x \backslash y}$
in $x \backslash y$ is all the belief in $x$ (of magnitude $b_x$) except for
that part which is also belief in $y$ (of magnitude $b_y$), so that 
$b_{x \backslash y} = b_x - b_y$.  The disbelief $d_{x \backslash y}$ in
$x \backslash y$ is equal to the sum of the disbelief in $x$ and the belief in
$y$.  The fact that $d_{x \backslash y} = d_x + b_y$ can be justified as
follows.  Since the state $y$ necessitates the
state $x$ ({\em i.e} the system must be in the state $x$ if it is in the
state $y$), then disbelief in $x \backslash y$ falls into two categories:
disbelief in $x$ (of magnitude $d_x$) and belief in $y$ (of magnitude $b_y$),
with the result that the disbelief in $x \backslash y$ should be $d_x + b_y$.

\subsection{Negation} 
\label{sec:negation}

\noindent
The negation of an opinion about proposition $x$ represents the
opinion about $x$ being false. This corresponds to `NOT' in binary
logic.

\begin{definition}[Negation]
\label{the:G12}
{  {
\hspace{1ex}\\
 Let ${\omega}_{{x}} =
(b_{x},d_{x},{u}_{x},{a}_{x})$ be an 
opinion about the proposition ${x}$.  Then ${\omega}_{\lnot{x}} =
(b_{\lnot{x}},d_{\lnot{x}},{u}_{\lnot{x}},{a}_{\lnot{x}})$ is the
negation of ${\omega}_{x}$ where:
\[
\begin{array}{l}
b_{\lnot{x}}= d_{x}\\
d_{\lnot{x}}= b_{x}\\
u_{\lnot{x}}= {u}_{x}\\
a_{\lnot{x}}= 1 - {a}_{x}\;.
\end{array}
\]
By using the
symbol `$\;\lnot$' to designate this operator, we define $\lnot
{\omega}_{x} \equiv {\omega}_{\lnot x}$. 
}  }
\end{definition}

Negation can be applied to expressions containing propositional
conjunction and disjunction, and it can be shown that De Morgans's
laws are valid.

\section{Products of Binary Frames of Discernment}
\label{prod-discern}

Multiplication and comultiplication in subjective logic are binary
operators that take opinions about two elements from distinct binary
frames of discernment as input parameters. The product and coproduct
opinions relate to subsets of the Cartesian product of the two binary
frames of discernment. The Cartesian product of the two binary frames
of discernment $X=\{x,\overline{x}\}$ and $Y=\{y,\overline{y}\}$
produces the quaternary set $X \times Y = \{(x,y),\; (x,\overline{y}),\;
(\overline{x},y),\; (\overline{x},\overline{y})\}$ which is illustrated in
Fig.\ref{fig:Cart-prod} below.

\begin{figure}[h]
\begin{center}
\includegraphics[scale=0.6]{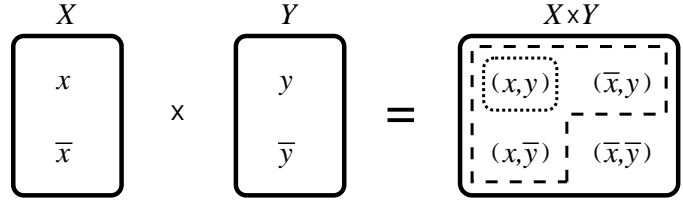}
\caption{Cartesian product of two binary frames of discernment}
\label{fig:Cart-prod}
\end{center}
\end{figure}

Let $\omega_{x}$ and $\omega_{y}$ be opinions about $x$ and $y$
respectively held by the same observer. Then the product opinion
$\omega_{x \land y}$ is the observer's opinion about the conjunction
$x \land y = \{(x,y)\}$ that is represented by the area inside the
dotted line in Fig.\ref{fig:Cart-prod}. The coproduct opinion
$\omega_{x \lor y}$ is the opinion about the disjunction $x \lor y =
\{(x,y),\; (x,\overline{y}),\; (\overline{x},y)\}$ that is represented
by the area inside the dashed line in Fig.\ref{fig:Cart-prod}.
Obviously $X \times Y$ is not binary, and coarsening is required in
order to determine the product and coproduct opinions. The reduced
powerset $2^{X \times Y}-\{\emptyset\}$ contains $2^{| X\times Y|}-1 =
15$ elements. A short notation for the elements of $2^{X \times Y}$ is
used below so that for example $\{(x,y),\; (x,\overline{y})\} = \{x\}
\times Y$. The bba on $X \times Y$ as a function of the opinions on
$x$ and $y$ is defined by:

\begin{equation}
\label{eq:cartesian-product}
\begin{array}{lll}
\begin{array}{r}
m(\{(x,y)\})                       = b_x b_y\;,\\
m(\{(x,\overline{y})\})            = b_x d_y\;,\\
m(\{x\} \times Y)                  = b_x u_y\;,\\
\end{array}
&\begin{array}{r}
m(\{(\overline{x},y)\})            = d_x b_y\;,\\
m(\{(\overline{x},\overline{y})\}) = d_x d_y\;,\\
m(\{\overline{x}\} \times Y)       = d_x u_y\;,\\
\end{array}
&\begin{array}{r}
m(X \times \{y\})                  = u_x b_y\;,\\
m(X \times \{\overline{y}\})       = u_x d_y\;,\\
m(X \times Y)                      = u_x u_y\;.
\end{array}
\end{array}
\end{equation}

It can be shown that the sum of the above belief masses always equals
1. The product does not produce any belief mass on the following
elements:

\begin{equation}
\begin{array}{lll}
\begin{array}{c}
\{(x,y),\; (\overline{x},\overline{y})\}\;,\\
\{(x,\overline{y}),\; (\overline{x},y)\}\;,
\end{array}\;\;\;
\begin{array}{c}
(X \times \{y\}) \cup \{(x,\overline{y})\}\;,\\
(X \times \{y\}) \cup \{(\overline{x},\overline{y})\}\;,
\end{array}\;\;\;
\begin{array}{c}
(X \times \{\overline{y}\}) \cup \{(x,y)\}\;,\\
(X \times \{\overline{y}\}) \cup \{(\overline{x},y)\})\;.
\end{array}
\end{array}
\end{equation}

The belief functions in for example $x \land y$ and $x \lor y$ can
now be determined so that:

\begin{equation}
\begin{array}{ll}
b(x \land y) = &m(\{(x,y)\}),\\
b(x \lor y)  = &m(\{(x,y)\})+m(\{(x,\overline{y})\})+m(\{(\overline{x},y)\})\;+\\
               &m(\{x\} \times Y) + m(X \times \{y\})\;.
\end{array}
\end{equation}

The normal base rate functions for $x \land y$ and $x \lor y$
can be determined by working in the respective ``primitive'' frames of
discernment, $\Theta_X$ and $\Theta_Y$ which underlie the definitions
of the sets $x$ and $y$, respectively.  A sample yields a value of
$(x,y)$ in the frame of discernment $X \times Y$ exactly when the
sample yields an atom $\theta_{X} \in x$ in the frame of discernment
$\Theta_X$ and an atom $\theta_{Y} \in y$ in the frame of discernment
$\Theta_Y$.  In other words, a sample yields a value of $(x,y)$ in the
frame of discernment $X \times Y$ exactly when the sample yields an
atom $(\theta_{X},\theta_{Y}) \in x \times y$ in the frame of discernment $\Theta_X
\times \Theta_Y$, so that $(x,y) \in X \times Y$ corresponds to $x
\times y \subseteq \Theta_X \times \Theta_Y$ in a natural manner.
Similarly, $(x,\overline{y})$ corresponds to $x \times \overline{y}$,
$(\overline{x},y)$ corresponds to $\overline{x} \times y$, and
$(\overline{x},\overline{y})$ corresponds to $\overline{x} \times
\overline{y}$.  The normal base rate function for $x \land y$
is equal to:
\begin{equation}
a(x \land y) = {{|x \times y|} \over {|\Theta_X \times \Theta_Y|}}
= {{|x|\; |y|} \over {|\Theta_X|\; |\Theta_Y|}} = {{|x|} \over {|\Theta_X|}}
\; {{|y|} \over {|\Theta_Y|}} = a(x) a(y),
\end{equation}
Similarly, the normal base rate of $x \lor y$ is equal to
\begin{eqnarray*}
a(x \lor y) & = & \frac{|(x \times y) \cup (x \times \overline{y}) \cup 
(\overline{x} \times y)|}{|\Theta_X \times \Theta_Y|}
= \frac{|x \times y| + |x \times \overline{y}| + |\overline{x} \times y|}
{|\Theta_X \times \Theta_Y|}\\
& = & \frac{|x|\; |y| + |x|\; |\overline{y}| + |\overline{x}|\; |y|}
{|\Theta_X|\; |\Theta_Y|}
= a(x) a(y) + a(x) a(\overline{y}) + a(\overline{x}) a(y)\\
& = & a(x) + a(y) - a(x) a(y).
\end{eqnarray*}

By applying simple or normal coarsening to the product frame of
discernment and bba, the normal product and coproduct
opinions emerge. A coarsening that focuses on $x \land y$ produces the
product, and a coarsening that focuses on $x \lor y$ produces the
coproduct. A Bayesian coarsening (i.e. when simple and normal coarsening
are equivalent) is only possible in exceptional cases because some
terms of Eq.(\ref{eq:cartesian-product}) other than $m(X \times Y)$
will in general contribute to uncertainty about $x \land y$ in the
case of multiplication, and to uncertainty about $x \lor y$ in the
case of comultiplication. Specifically, Bayesian coarsening requires $m(X
\times \{y\}) = m(\{x\}\times Y) = 0$ in case of multiplication, and
$m(X \times \{\overline{y}\}) = m(\{\overline{x}\}\times Y) = 0$ in
case of comultiplication. Non-Bayesian coarsenings will cause the product
and coproduct of opinions to deviate from the analytically correct
product and coproduct. However, the magnitude of this deviation is
always small, as shown in \cite{JM2004-IJAR}.

The symbols ``$\cdot$'' and ``$\sqcup$'' will be used to denote multiplication and comultiplication of opinions respectively so that we can write:

\begin{alignat}{2}
\omega_{x \land y} &\triangleq \omega_{x} \cdot \omega_{y}\\
\omega_{x \lor y} &\triangleq \omega_{x} \sqcup \omega_{y}
\end{alignat}

The product of the opinions about $x$ and $y$ is thus the opinion
about the conjunction of $x$ and $y$. Similarly, the coproduct of the
opinions about $x$ and $y$ is the opinion about the disjunction of $x$
and $y$. The exact expressions for product and coproduct are given in
Sec.\ref{normal-multiplication}.

Readers might have noticed that Eq.(\ref{eq:cartesian-product}) can
appear to be a direct application of the non-normalised version of
Dempster's rule (i.e. the conjunctive rule of combination)
\cite{Sha1976} which is a method of belief fusion. However the
difference is that Dempster's rule applies to the beliefs of two
different and independent observers faced with the same frame of
discernment, whereas the Cartesian product of
Eq.(\ref{eq:cartesian-product}) applies to the beliefs of the same
observer faced with two different and independent frames of
discernment. Let $\omega^{A}_{x}$ and $\omega^{B}_{x}$ represent the
opinions of two observers $A$ and $B$ about the same proposition $x$,
and let $\omega^{A,B}_{x}$ represent the fusion of $A$ and $B$'s
opinions. Let further $\omega^{A}_{x}$ and $\omega^{A}_{y}$ represent
observer $A$'s opinions about the propositions $x$ and $y$, and let
$\omega^{A}_{x \land y}$ represent the product of those
opinions. Fig.\ref{fig:fusion-product-difference} below illustrates
the difference between belief fusion and belief product.

\begin{figure}[h]
\begin{center}
\mbox{\subfigure[Belief fusion.]{\includegraphics[width=0.45\textwidth]{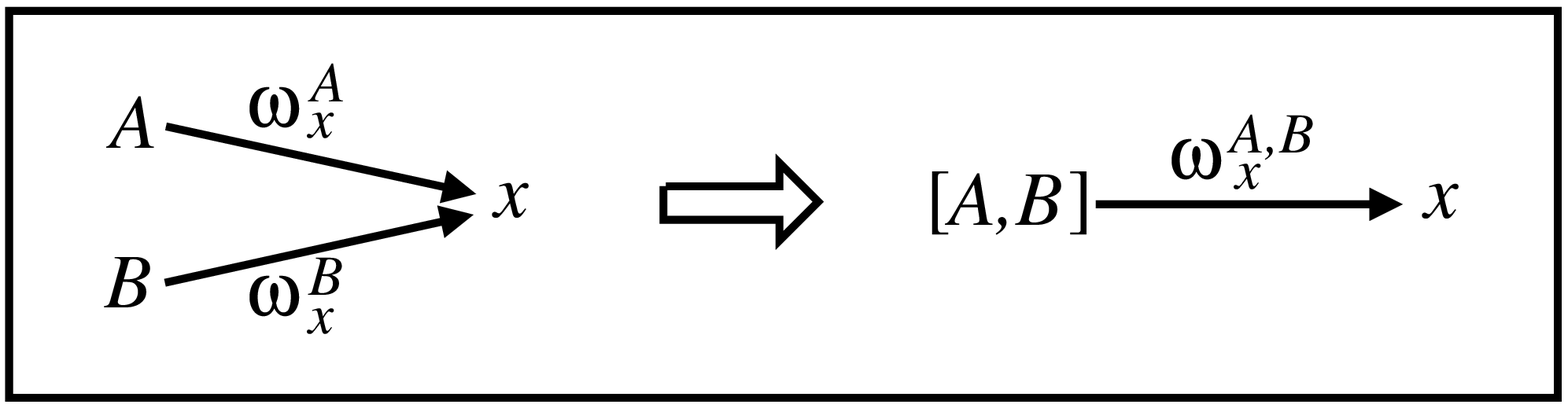}}
\quad
      \hspace{1mm}
      \subfigure[Belief product.]{\includegraphics[width=0.45\textwidth]{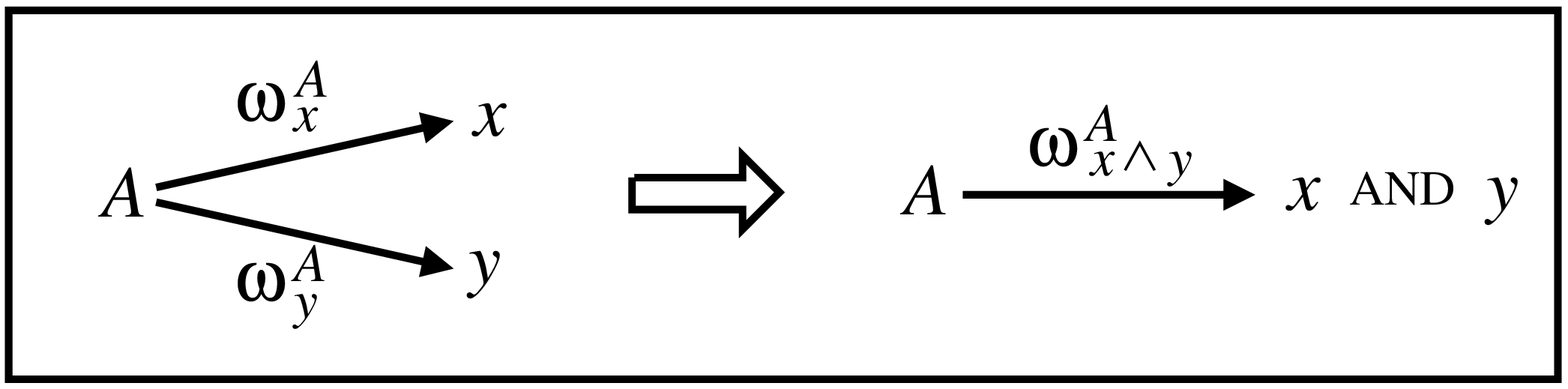}}}
\caption{Conceptual difference between belief fusion and belief product.}
\label{fig:fusion-product-difference}
\end{center}
\end{figure}

The Cartesian product as described here thus has no relationship to
Dempster's rule and belief fusion other than the apparent similarity
between Eq.(\ref{eq:cartesian-product}) and Dempster's rule.

\subsection{Normal Multiplication and Comultiplication}
\label{normal-multiplication}

Normal multiplication and comultiplication of opinions about independent
propositions $x$ and $y$ are based on normal coarsening defined in
\cite{JM2004-IJAR}. It is also straightforward to define
multiplication and comultiplication based on smooth coarsening, as
defined here. 

By the arguments within Section \ref{prod-discern} for
justifying the base rates, we can set $a_{x \land y} = a_x
a_y$ and $a_{x \lor y} = a_x + a_y - a_x a_y$.  This is in contrast to
the case of ``simple'' conjunction and ``simple'' disjunction as
discussed above, where atomicities of both the conjunction and the
disjunction are dependent on the beliefs, disbeliefs and uncertainties
of $x$ and $y$.  Given opinions about independent propositions, $x$
and $y$, then under normal coarsening of the bba for the Cartesian
product of the binary frames of discernment, the normal opinion for
the conjunction, $x \land y$, is given by
\begin{eqnarray*}
b_{x \land y} &=& \frac{(b_x + a_x u_x)(b_y + a_y u_y) - (1 - d_x)(1 - d_y)
a_x a_y}{1 - a_x a_y}\\
&=& b_x b_y + \frac{(1 - a_x) a_y b_x u_y + a_x (1 - a_y) u_x b_y}{1 - a_x 
a_y},\\
d_{x \land y} &=& d_x + d_y - d_x d_y,\\
u_{x \land y} &=& \frac{(1 - d_x)(1 - d_y) - (b_x + a_x u_x)(b_y + a_y u_y)}
{1 - a_x a_y}\\
&=& u_x u_y + \frac{(1 - a_y) b_x u_y + (1 - a_x) u_x b_y}
{1 - a_x a_y},\\
a_{x \land y} &=& a_x a_y.
\end{eqnarray*}

A numerical example of the normal multiplication operator is
visualised in Fig.\ref{fig:normalmultidemo} below. Note that in this
case, the base rate $a_{x \land y}$ is equal to the real
relative cardinality of $x \land y$ in $X \times Y$.

\begin{figure}[h]
\begin{center}
\includegraphics[angle=0,scale=0.630]{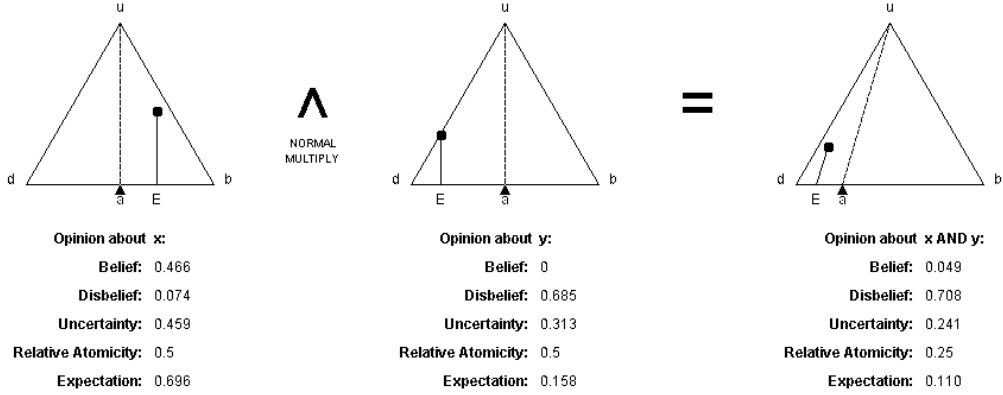}
\caption{Visualisation of numerical example of the normal multiplication operator}
\label{fig:normalmultidemo}
\end{center}
\end{figure}

The formulae for the opinion about $x \land y$ are well formed unless 
$a_x = 1$ and $a_y = 1$, in which case the opinions
$\omega_x$ and $\omega_y$ can be regarded as limiting values, and the product
is determined by the relative rates of approach of $a_x$ and $a_y$ to $1$.
Specifically, if $\eta$ is the limit of $\frac{1-a_x}{1-a_y}$, then
\begin{eqnarray*}
b_{x \land y} &=& b_x b_y + \frac{\eta b_x u_y + u_x b_y}{\eta + 1},\\
d_{x \land y} &=& d_x + d_y - d_x d_y,\\
u_{x \land y} &=& u_x u_y + \frac{b_x u_y + \eta u_x b_y}{\eta + 1},\\
a_{x \land y} &=& 1.
\end{eqnarray*}
Under normal coarsening of the bba for the Cartesian product of the binary
frames of discernment, the normal opinion for the disjunction, $x \lor y$, is
given by
\begin{eqnarray*}
b_{x \lor y} &=& b_x + b_y - b_x b_y,\\
d_{x \lor y} &=& \frac{(1 - (b_x + a_x u_x))(1 - (b_y + a_y u_y)) 
- (1 - b_x)(1 - b_y)(1 - a_x)(1 - a_y)}{a_x + a_y - a_x a_y}\\
&=& \frac{(d_x + (1 - a_x) u_x)(d_y + (1 - a_y) u_y) - (1 - b_x)(1 - b_y)
(1 - a_x)(1 - a_y)}{a_x + a_y - a_x a_y}\\
&=& d_x d_y + \frac{a_x (1 - a_y) d_x u_y + (1 - a_x) a_y u_x d_y}{a_x + a_y 
- a_x a_y},\\
u_{x \lor y} &=& \frac{(1 - b_x)(1 - b_y) - (1 - (b_x + a_x u_x))
(1 - (b_y + a_y u_y))}{a_x + a_y - a_x a_y}\\
&=& \frac{(1 - b_x)(1 - b_y) - (d_x + (1 - a_x) u_x)(d_y + (1 - a_y) u_y)}
{a_x + a_y - a_x a_y}\\
&=& u_x u_y + \frac{a_y d_x u_y + a_x u_x d_y}{a_x + a_y - a_x a_y},\\
a_{x \lor y} &=& a_x + a_y - a_x a_y.
\end{eqnarray*}

A numerical example of the normal comultiplication operator is
visualised in Fig.\ref{fig:normalcomultidemo} below. Note that in
this case, the base rate $a_{x \lor y}$ is equal to the real
relative cardinality of $x \lor y$ in $X \times Y$.

\begin{figure}[h]
\begin{center}
\includegraphics[angle=0,scale=0.630]{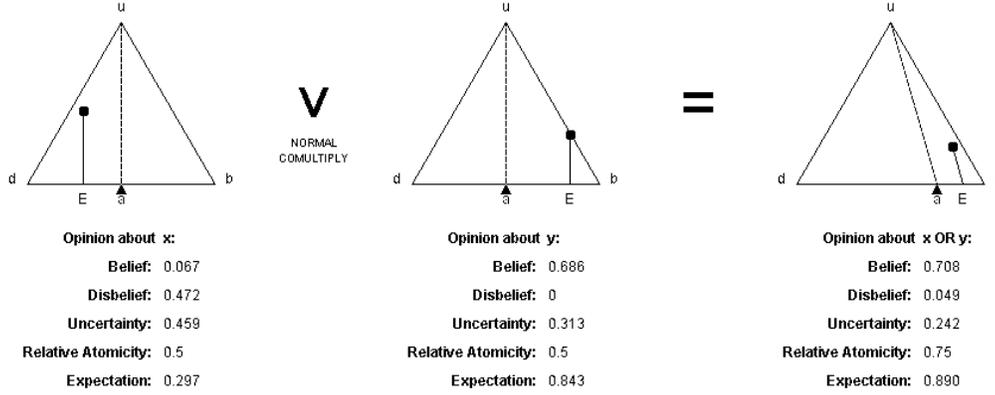}
\caption{Visualisation of numerical example of the normal comultiplication operator}
\label{fig:normalcomultidemo}
 \end{center}
\end{figure}

The formulae for the opinion about $x \lor y$ are well formed unless 
$a_x = 0$ and $a_y = 0$.  In the case that $a_x = a_y = 0$, the opinions
$\omega_x$ and $\omega_y$ can be regarded as limiting values, and the product
is determined by the relative rates of approach of $a_x$ and $a_y$ to $0$.
Specifically, if $\zeta$ is the limit of $\frac{a_x}{a_y}$, then
\begin{eqnarray*}
b_{x \lor y} &=& b_x + b_y - b_x b_y,\\
d_{x \lor y} &=& d_x d_y + \frac{\zeta d_x u_y + u_x d_y}{\zeta + 1},\\
u_{x \lor y} &=& u_x u_y + \frac{d_x u_y + \zeta u_x d_y}{\zeta + 1},\\
a_{x \lor y} &=& 0.
\end{eqnarray*}

This is a self-dual system under $b \leftrightarrow d$, $u \leftrightarrow u$,
$a \leftrightarrow 1 - a$, and $\land \leftrightarrow \lor$, that is, 
for example, the expressions for $b_{x \land y}$ and $d_{x \lor y}$ are dual 
to each other, and one determines the other by the correspondence, and 
similarly for the other expressions.  This is equivalent to the observation
that the opinions satisfy de Morgan's Laws, {\em i.e.}
$\omega_{x \land y} = \omega_{\overline{\overline{x} \lor \overline{y}}}$ and
$\omega_{x \lor y} = \omega_{\overline{\overline{x} \land \overline{y}}}$.

However it should be noted that multiplication and comultiplication are not distributive over each other, {\em i.e.} for example that:

\begin{equation}
\omega_{x \land (y \lor z)} \neq \omega_{(x \land y) \lor (x \land z)} 
\end{equation}

This is to be expected because if $x$, $y$ and $z$ are independent, then
$x \land y$ and $x \land z$ are not generally independent in probability
calculus. In fact the corresponding result only holds for binary logic.

\subsection{Normal Division and Codivision}
\label{normal-division}

The inverse operation to multiplication is division.  The quotient of
opinions about propositions $x$ and $y$ represents the opinion about a
proposition $z$ which is independent of $y$ such that $\omega_x =
\omega_{y \land z}$.  This requires that $a_x \leq a_y$, $d_x \geq
d_y$, and
\begin{eqnarray*}
b_x &\geq& \frac{a_x (1 - a_y) (1 - d_x) b_y}{(1 - a_x) a_y (1 - d_y)},\\
u_x &\geq& \frac{(1 - a_y) (1 - d_x) u_y}{(1 - a_x) (1 - d_y)}.
\end{eqnarray*}
The opinion $(b_{x \overline{\land} y},d_{x \overline{\land} y},
u_{x \overline{\land} y},a_{x \overline{\land} y})$, which is the
quotient of the opinion about $x$ and the opinion about $y$, is given by
\begin{eqnarray*}
b_{x \overline{\land} y} &=& \frac{a_y (b_x + a_x u_x)}{(a_y - a_x) 
(b_y + a_y u_y)} - \frac{a_x (1 - d_x)}{(a_y - a_x) (1 - d_y)},\\
d_{x \overline{\land} y} &=& \frac{d_x - d_y}{1 - d_y },\\
u_{x \overline{\land} y} &=& \frac{a_y (1 - d_x)}{(a_y - a_x) (1 - d_y)}
- \frac{a_y (b_x + a_x u_x)}{(a_y - a_x) (b_y + a_y u_y)},\\
a_{x \overline{\land} y} &=& \frac{a_x}{a_y},
\end{eqnarray*}
if $a_x < a_y$.  If $0 < a_x = a_y$, then the conditions required so that
the opinion about $x$ can be divided by the opinion about $y$ are
\begin{eqnarray*}
b_x &=& \frac{(1 - d_x) b_y}{1 - d_y},\\
u_x &=& \frac{(1 - d_x) u_y}{1 - d_y},
\end{eqnarray*}
and in this case,
\begin{eqnarray*}
d_{x \overline{\land} y} &=& \frac{d_x - d_y}{1 - d_y },\\
a_{x \overline{\land} y} &=& 1.
\end{eqnarray*}
The only information available about $b_{x \overline{\land} y}$ and
$u_{x \overline{\land} y}$ is that
\begin{eqnarray*}
b_{x \overline{\land} y} + u_{x \overline{\land} y} = 
\frac{1 - d_x}{1 - d_y}.
\end{eqnarray*}
On the other hand, $b_{x \overline{\land} y}$ and $u_{x \overline{\land} y}$
can be determined if the opinion about $x$ is considered as the limiting 
value of other opinions which can be divided by the opinion about $y$.  The 
limiting value of the quotient of the opinions is determined by the relative 
rates of approach of $a_x$, $b_x$ and $u_x$ to their limits.  Specifically,
if $\gamma$ is the limit of 
\begin{eqnarray*}
\frac{a_y (1 - a_y)}{(a_y - a_x) (b_y + a_y u_y)} \left(\frac{(1 - d_y) b_x}
{1 - d_x} - b_y\right) + \frac{b_y}{b_y + a_y u_y},
\end{eqnarray*}
then $0 \leq \gamma \leq 1$, and the limiting values of 
$b_{x \overline{\land} y}$ and $u_{x \overline{\land} y}$ are
\begin{eqnarray*}
b_{x \overline{\land} y} &=& \frac{\gamma (1 - d_x)}{1 - d_y},\\
u_{x \overline{\land} y} &=& \frac{(1 - \gamma) (1 - d_x)}{1 - d_y}.
\end{eqnarray*}

The inverse operation to comultiplication is codivision.  The co-quotient of 
opinions about propositions $x$ and $y$ represents the opinion about a 
proposition $z$ which is independent of $y$ such that $\omega_x = 
\omega_{y \lor z}$.  This requires that $a_x \geq a_y$, $b_x \geq b_y$, and 
\begin{eqnarray*}
d_x &\geq& \frac{(1 - a_x) a_y (1 - b_x) d_y}{a_x (1 - a_y) (1 - b_y)},\\
u_x &\geq& \frac{a_y (1 - b_x) u_y}{a_x (1 - b_y)}.
\end{eqnarray*}
The opinion $(b_{x \overline{\lor} y},d_{x \overline{\lor} y},
u_{x \overline{\lor} y},a_{x \overline{\lor} y})$, which is the
co-quotient of the opinion about $x$ and the opinion about $y$, is given by
\begin{eqnarray*}
b_{x \overline{\lor} y} &=& \frac{b_x - b_y}{1 - b_y },\\
d_{x \overline{\lor} y} &=& \frac{(1 - a_y) (1 - (b_x + a_x u_x))}
{(a_x - a_y) (1 - (b_y + a_y u_y))} - \frac{(1 - a_x) (1 - b_x)}{(a_x - a_y)
(1 - b_y)}\\
&=& \frac{(1 - a_y) (d_x + (1 - a_x) u_x)}{(a_x - a_y) (d_y + (1 - a_y) u_y)}
- \frac{(1 - a_x) (1 - b_x)}{(a_x - a_y) (1 - b_y)},\\
u_{x \overline{\lor} y} &=& \frac{(1 - a_y) (1 - b_x)}{(a_x - a_y) (1 - b_y)}
- \frac{(1 - a_y) (1 - (b_x + a_x u_x))}{(a_x - a_y) (1 - (b_y + a_y u_y))}\\
&=& \frac{(1 - a_y) (1 - b_x)}{(a_x - a_y) (1 - b_y)} - \frac{(1 - a_y)
(d_x + (1 - a_x) u_x)}{(a_x - a_y) (d_y + (1 - a_y) u_y)},\\
a_{x \overline{\lor} y} &=& \frac{a_x - a_y}{1 - a_y},
\end{eqnarray*}
if $a_x > a_y$.  If $a_x = a_y < 1$, then the conditions required so that
the opinion about $x$ can be codivided by the opinion about $y$ are
\begin{eqnarray*}
d_x &=& \frac{(1 - b_x) d_y}{1 - b_y},\\
u_x &=& \frac{(1 - b_x) u_y}{1 - b_y},
\end{eqnarray*}
and in this case,
\begin{eqnarray*}
b_{x \overline{\lor} y} &=& \frac{b_x - b_y}{1 - b_y },\\
a_{x \overline{\lor} y} &=& 0.
\end{eqnarray*}
The only information available about $d_{x \overline{\lor} y}$ and
$u_{x \overline{\lor} y}$ is that
\begin{eqnarray*}
d_{x \overline{\lor} y} + u_{x \overline{\lor} y} = \frac{1 - b_x}{1 - b_y}.
\end{eqnarray*}
On the other hand, $d_{x \overline{\lor} y}$ and $u_{x \overline{\lor} y}$
can be determined if the opinion about $x$ is considered as the limiting 
value of other opinions which can be codivided by the opinion about $y$.  The 
limiting value of the co-quotient of the opinions is determined by the relative 
rates of approach of $a_x$, $d_x$ and $u_x$ to their limits.  Specifically,
if $\delta$ is the limit of 
\begin{eqnarray*}
\frac{a_y (1 - a_y)}{(a_x - a_y) (d_y + (1 - a_y) u_y)}
\left(\frac{(1 - b_y) d_x}{1 - b_x} - d_y\right) + \frac{d_y}{d_y + (1 - a_y) u_y},
\end{eqnarray*}
then $0 \leq \delta \leq 1$, and the limiting values of 
$d_{x \overline{\lor} y}$ and $u_{x \overline{\lor} y}$ are
\begin{eqnarray*}
d_{x \overline{\lor} y} &=& \frac{\delta (1 - b_x)}{1 - b_y},\\
u_{x \overline{\lor} y} &=& \frac{(1 - \delta) (1 - b_x)}{1 - b_y}.
\end{eqnarray*}

Given the opinion about $x$ and the atomicity of $y$, it is possible
to use the triangular representation of the opinion space from
Fig.\ref{fig:tri-7-1} to describe geometrically the range of opinions
about $x \land y$ and $x \lor y$.

In the case of $x \land y$, take the projector for $\omega_x$, and
take the intersections of the projector with the line of zero
uncertainty and the line of zero belief ($A$ and $B$, respectively).
The intersection, $A$, with the line of zero uncertainty determines
the probability expectation value of $\omega_x$.  Take the point, $C$,
on the line of zero uncertainty whose distance from the disbelief
vertex is $a_y$ times the distance between the disbelief vertex and
$A$.  Take the line $BC$ and the line through $A$ parallel to $BC$.
Let $D$ and $E$ be the intersections of these lines with the line of
constant disbelief through $\omega_x$, so that the disbelief is equal
to $d_x$.  Then $\omega_{x \land y}$ falls in the closed triangle
determined by $D$, $E$ and the disbelief vertex, and the atomicity of
$x \land y$ is given by $a_{x \land y} = a_x a_y$.

\begin{figure}[h]
\begin{center}
\includegraphics[scale=0.50]{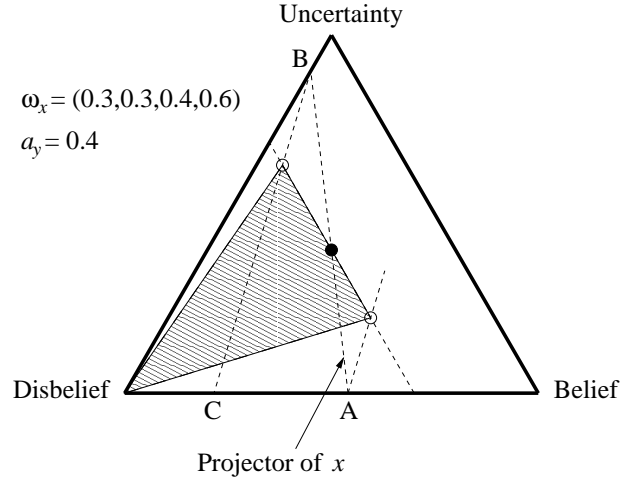}
\caption{Range of possible opinions for normal product}
\label{fig:tri-normal}
\end{center}
\end{figure}

In Fig. \ref{fig:tri-normal}, this is demonstrated with an example
where $\omega_x = (0.3, 0.3, 0.4, 0.6)$ and $a_y = 0.4$.  The opinion
$\omega_x$ has been marked with a small black circle (on the side of
the shaded triangle opposite the disbelief vertex).  The intersections $A$
and $B$ of the projector of $x$ with the line of zero uncertainty and
the line of zero belief, respectively, have been marked.  The point
$C$ has been placed on the probability axis so that its distance from
the disbelief vertex is 0.4 times the distance between $A$ and the disbelief 
vertex (since $a_y = 0.4$).  The line $BC$, whose direction
corresponds to an atomicity of 0.24 ({\em i.e.} the atomicity of $x
\land y$), has also been drawn in the triangle, and its intersection
with the line of constant disbelief through $\omega_x$ (with disbelief
equal to 0.3) has been marked with a white circle.  This is the point
$D$, although not marked as such in the figure.  The line through $A$
parallel to $BC$ has also been drawn in the triangle, and its
intersection with the line of constant disbelief through $\omega_x$
(the point $E$, although also not marked as such) has also been marked
with a white circle.  The triangle with vertices $D$, $E$ and the
disbelief vertex has been shaded, and the normal product $\omega_{x \land y}$
of the opinions must fall within the shaded triangle or on its
boundary.  In other words, the closure of the shaded triangle is the
range of all possible values for the opinion $\omega_{x \land y}$.

In the case of $x \lor y$, take the projector for $\omega_x$, and take
the intersections of this line with the line of zero uncertainty and
the line of zero disbelief ($A$ and $B$, respectively).  The
intersection, $A$, with the line of zero uncertainty determines the
probability expectation value of $\omega_x$.  Take the point, $C$, on
the line of zero uncertainty whose distance from the belief vertex is
$1 - a_y$ times the distance between the belief vertex and $A$.  Take
the line $BC$ and the line through $A$ parallel to $BC$.  Let $D$ and
$E$ be the intersections of these lines with the line of constant
belief through $\omega_x$, so that the belief is equal to $b_x$.  Then
$\omega_{x \lor y}$ falls in the closed triangle determined by $D$,
$E$ and the belief vertex, and the atomicity of $x \lor y$ is given by
$a_{x \lor y} = a_x + a_y - a_x a_y$.

The conditions required so that $\omega_x$ can be divided by $\omega_y$ can be 
described geometrically.  Take the projector for $\omega_y$, and take the 
intersections of this line with the line of zero uncertainty and with the line 
of zero belief.  Take the lines through each of these points which are parallel
to the projector for $\omega_x$ (it is required that $a_x < a_y$).  Take the 
intersections of these lines with the line of constant disbelief through 
$\omega_y$.  Then $\omega_x$ can be divided by $\omega_y$, provided $\omega_x$ 
falls in the closed triangle determined by these two points and the 
disbelief vertex.

Fig. \ref{fig:tri-normal} can be used to demonstrate.  If $\omega_y = (0.3,
0.3, 0.4, 0.6)$ and $a_x = 0.24$, then the black circle denotes $\omega_y$,
the projector of $y$ is the line $AB$, the lines through $A$ and $B$ parallel
to the director for atomicity 0.24 are drawn in the triangle, and their 
intersections with the line of constant disbelief through $\omega_y$ are 
marked by the white circles.  The closure of the shaded triangle is the
range of all possible values of $\omega_x$ that allow $\omega_x$ to be
divided by $\omega_y$.

The conditions required so that $\omega_x$ can be codivided by $\omega_y$ can 
be described geometrically.  Take the projector for $\omega_y$, and take the 
intersections of this line with the line of zero uncertainty and with the line 
of zero disbelief.  Take the lines through each of these points which are 
parallel to the projector for $\omega_x$ (it is required that $a_x > a_y$).  
Take the intersections of these lines with the line of constant belief through 
$\omega_y$.  Then $\omega_x$ can be codivided by $\omega_y$, provided 
$\omega_x$ falls in the closed triangle determined by these two points and 
the belief vertex.

\section{Principles of Belief Calculus}

Belief calculus follows the fundamental principles outlined below;
\begin{itemize}
\item The probability expectation value derived from an belief
expression, is always equal to the probability derived from the
corresponding probability expression. For example, let
$(\mbox{opinion expression})$ represent such a belief
expression, and let $(\mbox{probability expression})$ represent the
corresponding probability expression. By corresponding
expression is meant that every instance of a belief operator is
replaced by the corresponding probability operator,and the every
opinion argument is replaced by the probability expectation value of
the same opinion argument. Then we have:
\begin{equation}
\mathrm{E}(\mbox{opinion expression}) = (\mbox{probability expression})\;.
\end{equation}

\item The equivalence between opinions and augmented Beta density functions does not mean that the augmented Beta PDF that mapped from an opinion derived from a belief expressions is equal to the analytically correct probability density function. In fact, it is not clear whether it is possible to analytically derive probability density functions bases on algebraic expressions.

\item Because of the coarsening principles used in the (co)multiplication and (co)division operators, the opinions parameters can sometimes take illegal values, even when the argument opinions are legal. The general principle for dealing with this problem is to use clipping of the opinion parameters. This consists of adjusting the opinion parameters to legal values while maintaining the correct expectation value and base rate.

\item It is possible to design compact expressions for composite expressions, like e.g. conditional deduction as defined in \cite{JPD2005-ECSQARU}. The alternative is to use composite expressions as in Eq.\ref{eq:deduction}.

\end{itemize}

Belief calculus is based on the following fundamental belief operators.

\begin{table*}[h]
\begin{center}
\caption{Belief calculus operators. 
}
\vspace{2ex}
\begin{tabular}{rccl}
\hline
Opinion operator name &Opinion operator    &Logic operator  &Logic operator name\\
                      &notation              &notation  &\\
\hline
Addition              &$\omega_{x} + \omega_{y}$
                      &$x \cup y$   &UNION \rule[-3mm]{0mm}{8mm}\\
Subtraction           &$\omega_{x} - \omega_{y}$
                      &$x \backslash y$   &DIFFERENCE \rule[-3mm]{0mm}{8mm}\\
Multiplication        &$\omega_{x} \cdot \omega_{y}$
                      &$x \land y$   &AND \rule[-3mm]{0mm}{8mm}\\
Division              &$\omega_{x} / \omega_{y}$
                      &$x \overline{\land} y$ &UN-AND \rule[-3mm]{0mm}{8mm}\\
Comultiplication      &$\omega_{x} \sqcup\, \omega_{y}$   
                      &$x \lor y$    &OR \rule[-3mm]{0mm}{8mm}\\
Codivision            &$\omega_{x}\, \overline{\sqcup}\; \omega_{y}$
                      &$x \overline{\lor} y$ &UN-OR \rule[-3mm]{0mm}{8mm}\\
Complement            &$\lnot \omega_{x}$    &$\overline{x}$ &NOT \rule[-3mm]{0mm}{8mm}\\
\end{tabular}
\label{tab:operators}
\end{center}
\end{table*}

From these basic operators, a number of more complex operators can be
constructed. Some examples are for example:

\begin{itemize}
\item {\bf Conditional Deduction}. Input operands are the positive conditional $\omega_{y|x}$, the negative conditional $\omega_{y|\overline{x}}$ and the antecedent $\omega_{x}$ and its complement $\omega_{\overline{x}}$. From this, the consequent $\omega_{y\|x}$, expressed as:
\begin{equation}
\label{eq:deduction}
\omega_{y\|x} = \omega_{x} \cdot \omega_{y|x} + \omega_{\overline{x}} \cdot \omega_{y|\overline{x}}\;.
\end{equation}

\item {\bf Conditional Abduction}. Input operands are the positive
conditional $\omega_{x|y}$, the negative conditional
$\omega_{x|\overline{y}}$, the base rate of the consequent $y$, the
antecedent opinion $\omega_{x}$ and its complement
$\omega_{\overline{x}}$. This requires the computation of reverse
conditionals according to:

\begin{eqnarray}
\omega_{y|x} = \frac{\omega_{y} \cdot \omega_{x|y}}{\omega_{y} \cdot \omega_{x|y} + a_{\overline{x}} \cdot \omega_{x|\overline{y}}}\;.\\
\nonumber \\
\nonumber \\
\omega_{y|\overline{x}} = \frac{\omega_{y} \cdot \lnot \omega_{x|y}}{\omega_{y} \cdot \lnot \omega_{x|y} + a_{\overline{x}} \cdot \lnot \omega_{x|\overline{y}}}\;.
\end{eqnarray}

These two expressions make it possible to compute $\omega_{y\|x}$
according to Eq.\ref{eq:deduction}.
\end{itemize}

\section{Conclusion}

Belief calculus represents a general method for probability calculus
under uncertainty. The simple general operators can be combined in any
number of ways to produce composite belief expressions. Belief
calculus will always be consistent with traditional probability
calculus.

Belief calculus represents an approximative calculus for uncertain
probabilities that are equivalent to Beta probability density
functions. The advantage of this approach is that it provides an
efficient way of analysing models expressed in terms of Beta PDF which
otherwise would be impossible or exceedingly complicated to analyse.

\bibliographystyle{plain}
\bibliography{../bibliography}

\end{document}